\documentclass[10pt, final]{article}
\usepackage{multicol}
\usepackage{cite}
\usepackage{fullpage}
\usepackage{graphicx}
\usepackage{hyperref}
\usepackage{amsmath,float}
\usepackage{subcaption}
\usepackage{url}
\usepackage[ruled,vlined,commentsnumbered,titlenotnumbered,norelsize]{algorithm2e}
\usepackage{geometry}
\newcommand{\br}[1][.75]{\ \\[#1\baselineskip]}

\begin{document}
	\begin{center}
		\LARGE{\textbf{AI in Game Playing: Sokoban Solver}}\\
		\Large{\textbf{CS 221 Project Progress Report}}\\
		\Large{Anand Venkatesan, Atishay Jain, Rakesh Grewal }
	\end{center}
	\begin{multicols}{2}	
	\section{Introduction}
	Artificial Intelligence is becoming instrumental in a variety of applications. Games serve as a good breeding ground for trying and testing these algorithms in a sandbox with simpler constraints in comparison to real life.	In this project, we aim to develop an AI agent that can solve the classical Japanese game of Sokoban using various algorithms and heuristics
	and compare their performances through standard metrics. \br 
	In this progress report, we delve deeper into the ideas furnished in our proposal by (1) pointing out the game mechanics in a straightforward manner (2) describing the states and explaining how it is modeled in our algorithms (3) detailing our algorithms to calculate the best moves for the levels of the game (4) defining proper pruning techniques that are implemented to ameliorate the performance of the algorithm (5) providing results and comparing the developed algorithm using standard metrics of evaluation. We conclude with the next steps we plan to take up to finish this project.
	
	\section{General Game Mechanics}
Sokoban, literally referring to a warehouse keeper, was created by Hiroyuki Imabayashi and is a cult classic. This game is a transportation puzzle where the playing arena is composed of a grid of squares. Some of the squares are marked as crates where the player has to push to a storage location in the warehouse. Some of the squares are marked as walls which act as constraints where the player as well as the crates cannot enter. \br
The initial state consists of a player at a certain x,y location on the grid and certain locations marked as crates(or boxes) and target stores. The player can move horizontally or vertically (four directions - Up, Down,Left and Right). The player can push at most a single box into an empty space that is not a wall or another box. The player is not allowed to pull the boxes too. There are equal number of crates and target locations and the player succeeds once all crates are in target storage locations. The player fails if a crate gets locked up in a corner or with another crate with the storage locations(or location) being unoccupied.
	\section{Literature Review}
	From the literature study, we gather that the algorithm for solving this highly popular transportation game is researched extensively and various implementations have been formulated for quickly solving this problem with better efficiency. Previous study shows that many heuristics [2] like PI-corral pruning (which reduces the number of positions expanding), hashing (which avoids searching the same position multiple times), deadlock	tables (which calculates the positions of the deadlocks in the game) etc. are used to solve the game. Solvers	using BFS-A*, DFS-A* are implemented in [3] and are compared with the rudimentary and naive techniques. It is interesting to note that some authors have attempted to translate this problem into a machine learning problem where the machine is used to generate new levels based on the complexity desired by the user [4].
	\section{Scientific Value and Scope}
	Solving Sokoban is a NP-Hard problem, PSPACE-Complete [1] and it has been an active area of research. The branching factor of the Sokoban game is very high and with each iteration, it has an exponential number of pushes and moves. Therefore it needs proper heuristics that can help in eliminating redundant search states. The backtracking algorithm limitations are evident when the size of the puzzle is huge. Solving Sokoban has useful applications in robotics, especially motion planning [4]. The robotic movement in a	constrained space can be simplied to Sokoban.
	\section{States and Modeling}
	The game of sokoban can alternatively be considered as a search problem where we essentially look out for boxes and storage locations. So intuitively, it has a valid start state and end state. The start state is the state given by the original game developer whereas the end state is the state when all crates are transferred to proper storage locations. The actions can be moving in all directions with a cost associated with it which leads to the successor state. 
		\begin{center}
		\centering
		\includegraphics[width=5cm, height=5cm]{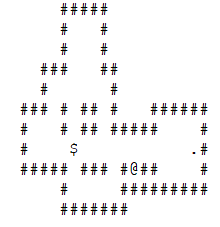}
		\captionof{figure}{Game model}
	\end{center}
	\begin{figure*}
	\centering
	\includegraphics[height=1.3in]{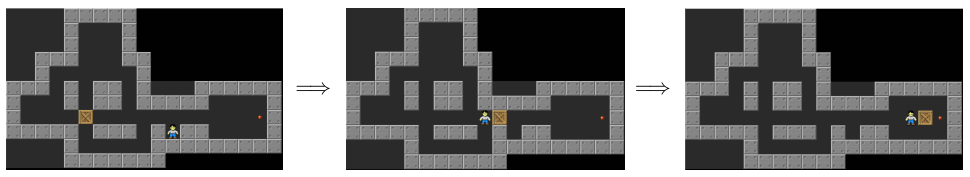}
	\caption{Game play of Sokoban}
	\label{fig:chain}
\end{figure*}
	For modeling this game, we have a standard notation that is used to independently and distinctly distinguish between all the objects in the game. Each level (\textit{like that in Figure 1}) is given in a unique representation. The inputs for modeling consist of the following characters: ``\#'', `` ", ``\$", ``@", ``." and ``+". Each of these characters have a special attribute of the game assigned to it: \textbf{``\#"} is a wall,  \textbf{`` "} is a free space, \textbf{``\$"} is a box, \textbf{``."} is a goal place,  \textbf{``*"} is a boxes placed on a goal, \textbf{``@"} is for Sokoban and \textbf{``+"} is for Sokoban on a goal. So the level described in Figure 1 is modified as follows:  \\
	\section{Algorithms}
	Sokoban has many unique properties which are not there in other similar problems as Rubik’s cube or Lloyd Fifteen Puzzle. Moves are irreversible so making a bad move can make you lose the game. Given the game mechanics, several factors are taken into account for deciding on an optimal algorithm:
	(1) There is a need for returning a set of moves quickly because the game is played real time also. Consequently, an algorithm that is able to produce moves in a short amount of processing time and returns a strong move is desirable.
	(2) The constraint added to game. Constraints in this game refers to the walls that are present inside the outer boundary and restrict the path of the player.
	(3) The depth to which the search is constructed. Since this search problem can lead to multitudes of states and can at times lead to infinite search depths, once concern is to know how much the algorithm can explore in the search space. \br
	Given these considerations, we decided to evaluate considerable number of algorithms and compare them based on their space/time complexities. The strength of the algorithms in this space lies in their ability to quickly determine sequences of moves that yield relatively strong results. We have implemented four algorithms namely backtracking, Depth First Search (DFS), Breadth First Search (BFS) and Uniform Cost Search (UCS) and shown their results.
	\subsection{Baseline (\textit{Backtracking Algorithm)} and Oracle Implementation}
	The baseline of the project is the backtracking search algorithm and the oracle of the project is the high level human intelligence that is used to solve the game. So essentially, each level has a predefined number of moves which corresponds to the minimum moves that one can take to solve the game. This minimum number of steps forms the oracle. On the other hand, the backtracking algorithm is one of the simplest recursive algorithms and forms the baseline for our project but is seldom used widely because of its high time complexity. It just recurses to all states and finds the minimum cost in reaching the goal. The space complexity is O(D) and the time complexity is O($b^D$) which is very high. 
	\subsection{Depth First Search}
	Depth First Search (DFS) is a special case of backtracking search algorithm. The search starts from the root and proceeds to the farthest node before backtracking. The difference between this and the backtracking is that this stops the search once a goal is reached and does not care if it is not minimum. The space and time complexities, on the worst case, are the same as the baseline algorithm but stops when it finds the solution. 
	\subsection{Depth First Search with Iterative Deepening}
	The Depth First Search with Iterative Deepening (DFS-ID) algorithm is a small addition to the DFS with a Maxdepth added to stop it. This increases the order of depths from one till maxdepth and typically performs the same action as that of DFS. 
	\subsection{Breadth First Search}
	Breadth First Search (BFS), as the name says, explores the search space in the increasing order of the depth and the costs of traveling from one state to another is assumed to be a positive number. Typically, this algorithm is often associated with the concept of stack and queue and pushing and popping from the stack. Due to the larger states explored at shorter depths, the space complexity is very high of about O($b^d$) and the time complexity is O($b^d$). 
	\subsection{Uniform cost Search}
	For any search problem, Uniform Cost Search (UCS) is the better algorithm than the previous ones. The search algorithm explores in branches with more or less same cost. This consist of a priority queue where the path from the root to the node is the stored element and the depth to a particular node acts as the priority. UCS assumes all the costs to be non negative. While the DFS algorithm gives maximum priority to maximum depth, this gives maximum priority to the minimum cumulative cost. 
	\subsection{A* Algorithm}
	A* algorithm is one of the popular technique used in path finding and graph traversals. This algorithm completely relies on heuristics for computing the future cost of a problem. This algorithm is equivalent to the uniform cost search with modified edge cost. This heuristics is chosen according to the case where the algorithm is implemented, thus emphasizing the importance of domain knowledge. This algorithm is consistent if the modified cost is greater than zero.
	\subsection{Convolutional Neural Network}
	With the search tree being exponential and the problem being NP-Hard, modeling and brute force inference can only go to a certain limit to solve a complex problem like Sokoban. We further explore learning by experimenting the convolutional neural network approach. For learning the game, we pass a representation of the state to a convolutional neural network and train it based on the best possible action for that state. The supervised learning approach generally performs faster than reinforcement learning especially in deterministic games like sokoban because the data used for training is significantly more.
		\begin{center}
		\includegraphics[width=8cm, height=5cm]{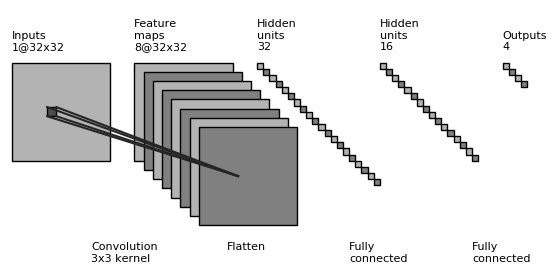}
		\captionof{figure}{Covolutional Neural Network}
	\end{center}
	\section{Pruning Techniques}
	Pruning is a terminology used in machine learning and artificial intelligence that are used to reduce the size of the decision trees by removing the selected sections of the tree that provide undesirable results. We have implemented two techniques till now in the project and are described below:
	\subsection{Move appension}
	One of the primary problem in search problems is that the computation time becomes unimaginably high when the search space is big. In such cases, if we have priori knowledge about the systems, we can limit in the beginning of the search problem all the cases where we have impossible actions. For instance, the Figure 3 depicts a case where the only acceptable action is to move Up. The possible actions for any algorithm can be moving in all the directions which can be reduced to one by Move Appension where we restrict all the impossible actions. 
	\subsection{Hashing}
	Hashing is a well known pruning method used to tune the algorithm to perform better. It follows the logic that decision which leads to the states that are already visited are considered as suboptimal. So all the states are stored in the hash table and at each point, a comparison is made between the current state and the stored state. If there is a match, the same action corresponding to the one in the hashing table is avoided.
		\begin{center}
		\includegraphics[width=6cm, height=5cm]{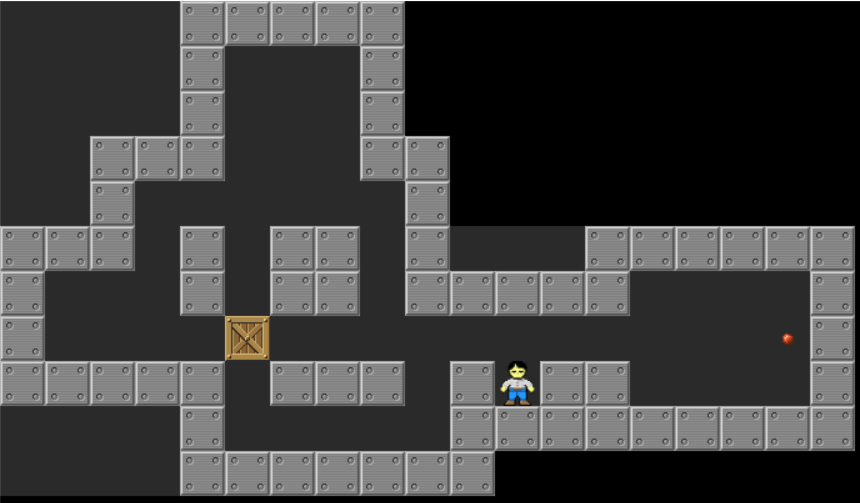}
		\captionof{figure}{Move Appension Example}
	\end{center}
	\subsection{Tunnel Macros}
	Tunnel Macros is a intelligent pruning technique which is employed to reduce the state space exploration. Very often, the Sokoban game consist of the tunnels where the boxes are required to be pushed. In such cases, the action is not going to change till the tunnel end is reached. Therefore, if we can cut down the actions in the tunnels by identifying them, the time required to solve the level decreases exponentially.  
		\begin{figure*}
		\centering
				\includegraphics[height=1.15in]{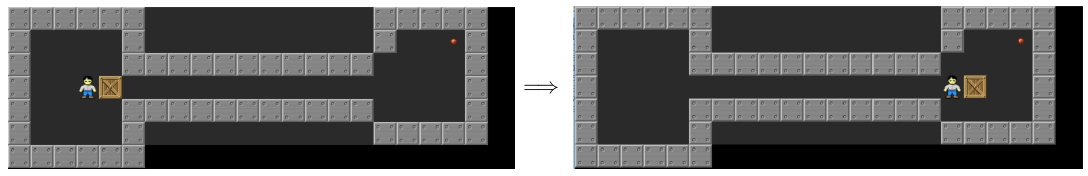}
		\caption{Tunnel Macros}
	\end{figure*}
	\section{Results}
	The performances of all the algorithm are calculated based on standard metrics like the time consumed, total number of states explored and the solution steps and are graphically represented in the \textbf{Figure 6}, \textbf{Figure 7} and \textbf{Figure 8}. The algorithms were implemented using two important characteristics. One was the time-out and the other was the Max-Depth. The algorithms were made to stop when it reached a certain predefined depth called the Max-Depth and is also terminated when it exceeded the maximum time, Time-out. This was necessary in order to prevent the algorithms from searching in infinite search space.\br
	It can be observed that the DFS exceeded the maxdepth and failed to perform in higher Levels. The time taken for it is significantly lesser than Backtracking but is greater than the other two algorithms. It was also observed that the states explored by all the algorithms were almost the same but they varied only by the time consumed and the solution steps.The oracle had a little higher performances than the algorithms. For instance, the number of steps taken for the Levels 1-5 were 33, 43, 57, 82, 51 respectively.
	There are two cost functions considered for UCS. The first cost function was implemented by giving maximum cost to the action of pushing the crate out of target location followed by moving the crates and the least cost is given to movement of the player. The second cost function was taken in a way where the first and third actions had equal cost. When this was implemented, the first cost function performed well whereas a sluggish actions were witnessed for the second cost function. 
				\begin{center}
		\includegraphics[width=8.5cm, height=5cm]{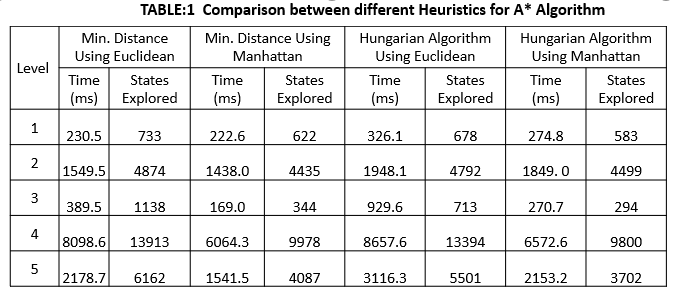}
	\end{center}  \br
	We considered four heuristics for A* Algorithm. The first heuristic was implemented by considering minimum distance computed by Euclidean distance between the targets and the boxes. The second heuristic was computed by calculating the minimum distance with Manhattan distance between them. The last two heuristic involved Hungarian logic with Euclidean and Manhattan distances where we considered the minimum distance between each target and the boxes and adding one more constraint that each box must be mapped to only one target. Their performances are tabulated in \textbf{Table 1}. It was observed that the Hungarian logic was considering lesser states of exploration because it mimicked the original constraint more than the other. On the other hand, the time taken by it is more than the Minimum Distance metric. 
		\begin{center}
		\includegraphics[width=9cm, height=4.5cm]{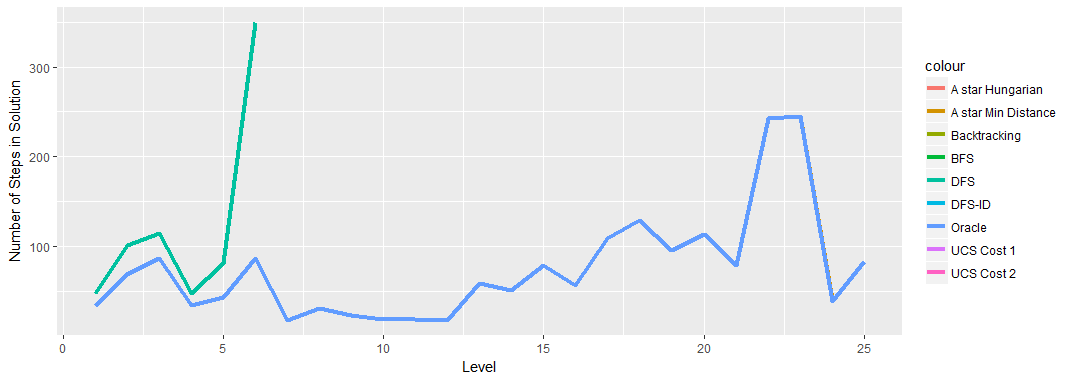}
		\captionof{figure}{Time Consumed and Solution Steps}
	\end{center}
	\begin{center}
		\includegraphics[width=9cm, height=4.5cm]{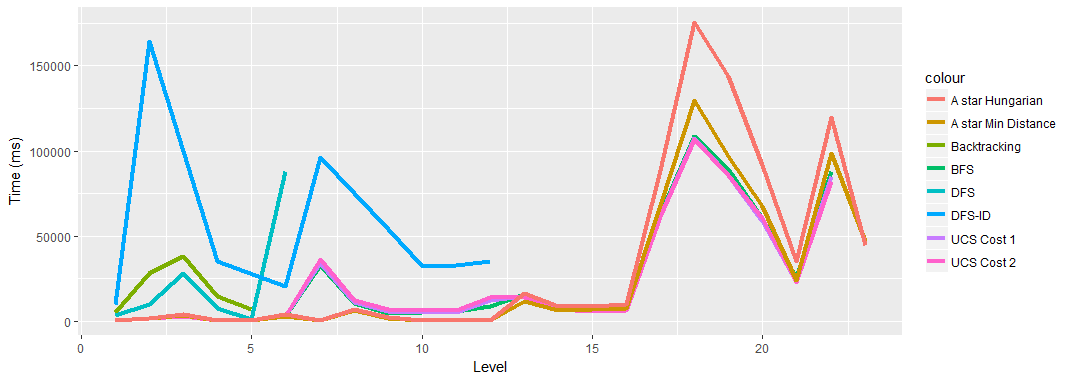}
		\captionof{figure}{Time Consumed and Solution Steps}
	\end{center}
	\begin{center}
	\includegraphics[width=9cm, height=4.5cm]{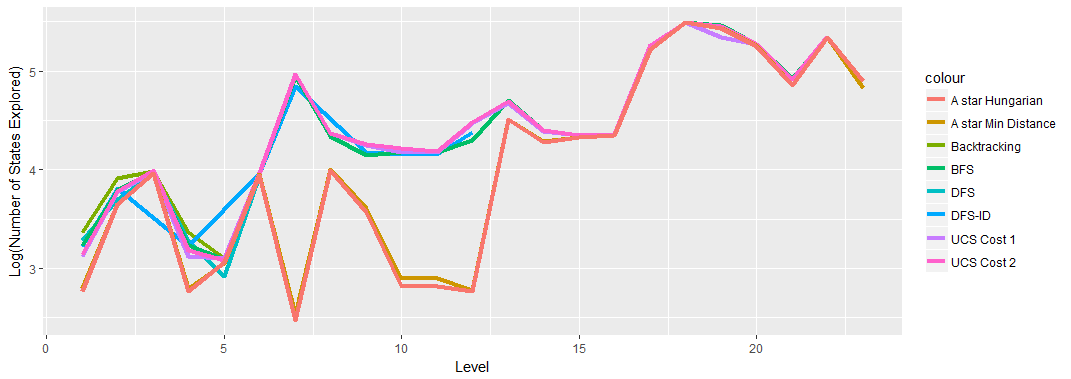}
	\captionof{figure}{Time Consumed and Solution Steps}
\end{center}
			\begin{center}
		\includegraphics[width=9cm, height=6.5cm]{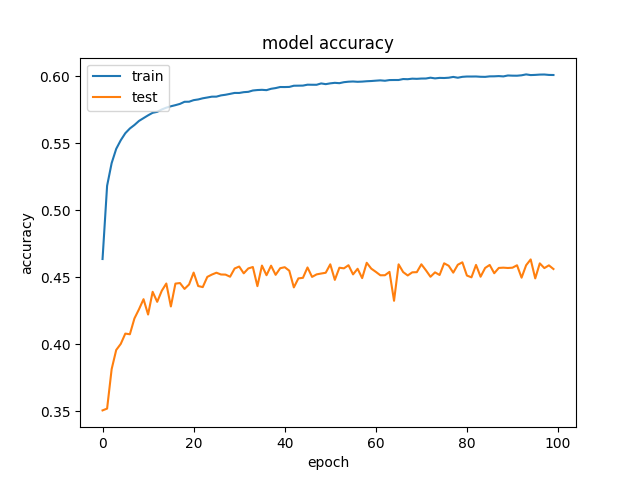}
		\captionof{figure}{Accuracy of CNN}
	\end{center}
	In our approach for CNN, we generated all possible states and their corresponding best actions using the backtracking algorithm for simple states and provided the state and the corresponding action to a convolutional neural network. 
	Our network consists of 8 3x3 convolutions over the single 32x32 representation of a state, followed by two sets of hidden layers of size 32 and 16  with all the values found empirically. The output layer with four possible actions classified over softmax. We used the relu non-linearity for our non-terminal layers. The dataset consists of $\approx$ 10000 states with a 10\% test data split. \br
			\begin{center}
		\includegraphics[width=8cm, height=6cm]{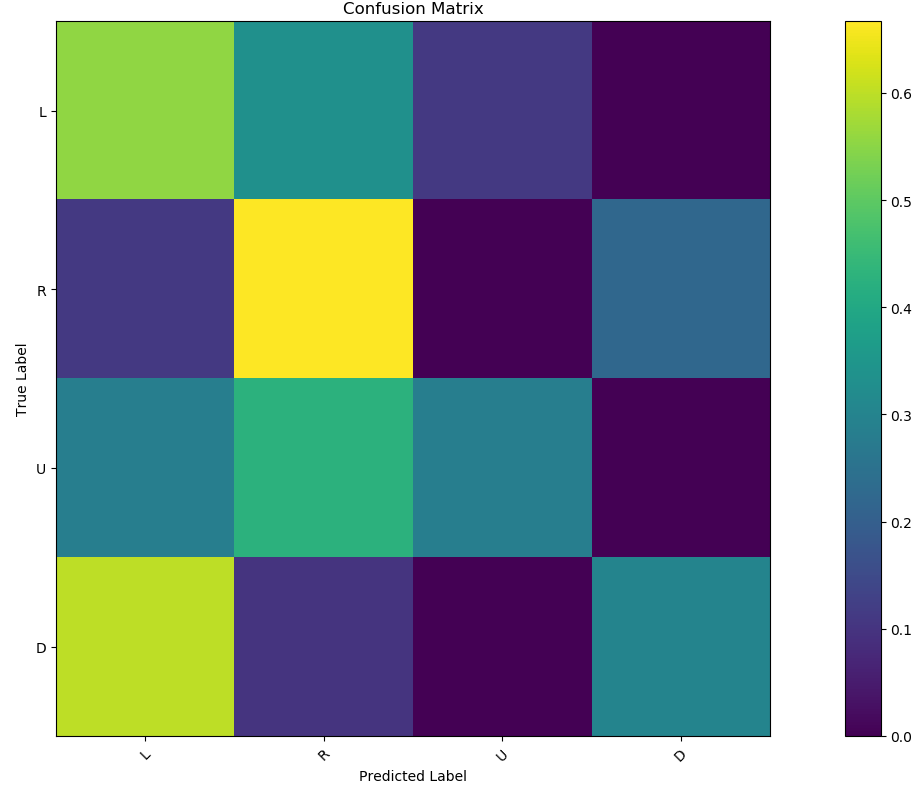}
		\captionof{figure}{Normalised Confusion Matrix for CNN}
		{Level 15}
	\end{center}
	We use data augmentation to increase our training data by using rotation and reflection of our levels to generate 8 times the original data gathered by the backtracking model. This also ensures that that all possible outputs(LRUD) have an equal 25\% probability and the level generation bias is removed from our training and test data.
	When playing the actual game with the trained model, it is observed that the sokoban falls into cycles and loops that had been removed by our pruning techniques in the search based sokoban solver. To force the game to finish, we take the next best option for any state where the model has reached previously.
	The model trains with 58\% train and 45\% test accuracy. This poor performance is due to the complex nature of the game. One incorrect move makes the user fail the level which is highly difficult to model using CNN. The positive side is that this algorithm performs in all the levels irrespective of the dimension of the level. The other algorithms take infinitely more time whereas this algorithm is instantaneous in its result. The confusion matrices for one of the levels is shown in \textbf{Figure 10}.
The algorithms for Backtracking, Depth First Search, Breadth First Search and Uniform Cost Search are given in the Appendix Section 1. The game play of the Sokoban game by the Uniform Cost Search Algorithm with the first cost function is implemented step by step and is provided in the Section 12.
	\section{Conclusion}
	We have implemented many algorithms for developing an AI agent for Sokoban game. We experimented with some of the pruning techniques and observed that most of the algorithms were having similar moves in playing the game but their performances differed in the time and the states explored. Given a smaller dimension of level, A* algorithm with Hungarian distance metric computed by Manhattan distance found to have better performance than the other algorithms considered. But given a larger dimensions of the level, all the algorithms are constrained by time and the Convolutional Neural Network had fairly decent performance.  
	\section{Next Steps}
	We plan to implement the following in the future:
	\begin{itemize}
		\item implementing dynamic deadlock technique to the database to check future positions.
		\item exploring and trying to implement level symmetries which can significantly improve the performance of the algorithms as most of the levels are symmetric in parts.
		\item trying to make the CNN work better by training it more accurately to achieve better performances.
	\end{itemize}
\section{Reference}
$[1]$ Dor, Dorit, and Uri Zwick. "SOKOBAN and other motion planning problems." Computational Geometry 13.4 (1999): 215-228.\\
$[2]$ http://pavel.klavik.cz/projekty/solver/solver.pdf \\
$[3]$ Li, Zheng, et al. "Object-oriented Sokoban solver: A serious game project for OOAD and AI education." Frontiers in Education Conference (FIE), 2014 IEEE. IEEE, 2014. \\
$[4]$ Taylor, Joshua, and Ian Parberry. "Procedural generation of Sokoban levels." Proceedings of the International North American Conference on Intelligent Games and Simulation. 2011.
\end{multicols}
\section{CodaLab Link}
\href{https://worksheets.codalab.org/worksheets/0x2412ae8944eb449db74ce9bc0b9463fe/}{https://worksheets.codalab.org/worksheets/0x2412ae8944eb449db74ce9bc0b9463fe/}
	\section{Algorithms}
	\begin{algorithm}\label{back}\small
	\caption{\small Backpropagation Algorithm$(state, maxdepth, maxtimeout)$}
	stack $\gets$ starting position of Sokoban \\
	\While{stack is not empty}{
		\eIf{Are crates on target}{put the state in option}{
			\eIf{Is deadlock or is depth $\ge$ maxdepth or is time $\ge$ maxtimeout}{pick next solution}{
				Get valid moves for Sokoban \\
				\ForEach{move}{
					Find next state \\
					Put it in stack	
			}}
	}}
	Pick the Best moves
\end{algorithm}

\begin{algorithm}\label{DFS}\small
	\caption{\small Depth First Search Algorithm$(state, maxdepth, maxtimeout)$}
	stack $\gets$ starting position of Sokoban \\
	\While{stack is not empty}{
		\eIf{Are crates on target}{break}{
			\eIf{Is deadlock or is depth $\ge$ maxdepth or is time $\ge$ maxtimeout}{pick next solution}{
				Get valid moves for Sokoban \\
				\ForEach{move}{
					Find next state \\
					Put it in stack	
			}}
	}} 
	return moves
\end{algorithm}

\begin{algorithm}\label{DFS-ID}\small
	\caption{\small Depth First Search with Iterative Deepening Algorithm$(state, maxdepth, maxtimeout)$}
	stack $\gets$ starting position of Sokoban \\
	depth $\gets$ 1
	\While{stack is not empty}{
		\eIf{Are crates on target}{break}{
			\eIf{Is deadlock or is time $\ge$ maxtimeout}{increase the depth \\
				pick next solution}{
				Get valid moves for Sokoban \\
				\ForEach{move}{
					Find next state \\
					Put it in stack	
			}}
	}} 
	return moves
\end{algorithm}

\begin{algorithm}\label{BFS}\small
	\caption{\small Breadth First Search Algorithm$(state, maxdepth, maxtimeout)$}
	queue $\gets$ starting position of Sokoban \\
	cost $\gets$ cost of moves \\
	\While{queue is not empty}{
		Remove the first element of queue
		\eIf{Are crates on target}{break}{
			\eIf{Is deadlock or is depth $\ge$ maxdepth or is time $\ge$ maxtimeout}{pick next solution}{
				Get valid moves for Sokoban \\
				\ForEach{move}{
					Find next state \\
					Put it in queue with current cost+1	
			}}
	}} 
	return moves
\end{algorithm}

\begin{algorithm}\label{UCS}\small
	\caption{\small Uniform Cost Search Algorithm$(state, maxtimeout)$}
	priority queue $\gets$ starting position of Sokoban \\
	cost $\gets$ cost of moves \\
	\While{queue is not empty}{
		Remove the highest priority element of queue
		\eIf{Are crates on target}{break}{
			\eIf{Is deadlock or is depth $\ge$ maxdepth or is time $\ge$ maxtimeout}{pick next solution}{
				Get valid moves for Sokoban \\
				\ForEach{move}{
					cost $\gets$ cost of move\\
					Add cost to current move
					Update queue with new cost and new state	
			}}
	}} 
	return moves
\end{algorithm}

\begin{algorithm}\label{A*}\small
	\caption{\small A* Algorithm$(state, maxdepth, maxtimeout)$}
	priority queue $\gets$ starting position of Sokoban \\
	cost $\gets$ cost of moves \\
	Fix the Heuristics
	\While{queue is not empty}{
		Remove the highest priority element of queue
		\eIf{Are crates on target}{break}{
			\eIf{Is deadlock or is depth $\ge$ maxdepth or is time $\ge$ maxtimeout}{pick next solution}{
				Get valid moves for Sokoban \\
				\ForEach{move}{
					cost $\gets$ cost of move with heuristics\\
					Add cost to current move
					Update queue with new cost and new state	
					Find Heuristics of new state
			}}
	}} 
	return moves
\end{algorithm}

\begin{algorithm}\label{CNN}\small
	\caption{\small Convolutional Neural Network$(state, maxdepth, maxiter, maxtimeout)$}
	Generate Levels by rotation and mirroring \\
	Run Backtracking Algorithm for all levels with no time/space constraints\\
	Fix ytrue $\gets$ Result of Backtracking\\
	x $\gets$ states \\
	\While{iteration $\le$ maxiter}{
		Fit CNN Model using Train Levels\\
		Update weights by SGD\\
	}
	Save the parameters \\
	Run the Test Levels \\
	Compute Accuracy 
\end{algorithm}

\newpage
	\section{Glimpse of Solution for Level 3 using A* Algorithm}
		\begin{figure}[h!]
		{
			\begin{subfigure}[h!]{1.3in}
				\includegraphics[height=1.5in]{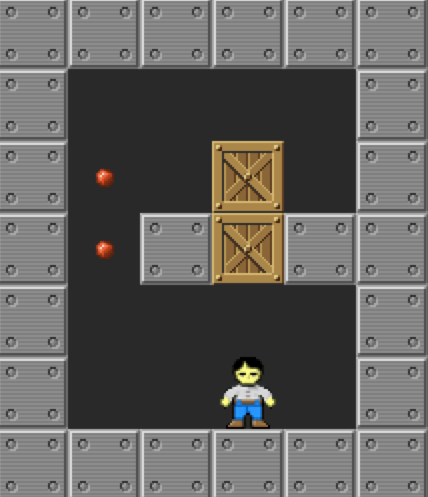}
			\end{subfigure}
			\begin{subfigure}[h!]{1.3in}
				\includegraphics[height=1.5in]{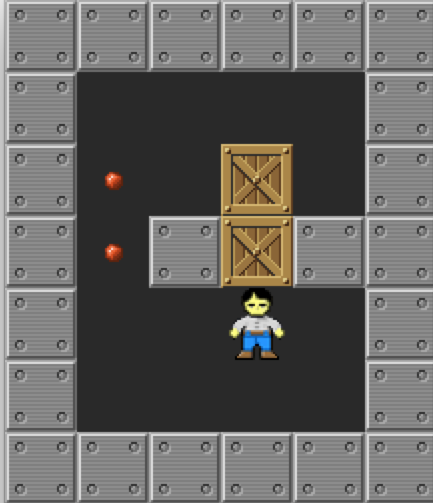}
			\end{subfigure}
			\begin{subfigure}[h!]{1.3in}
				\includegraphics[height=1.5in]{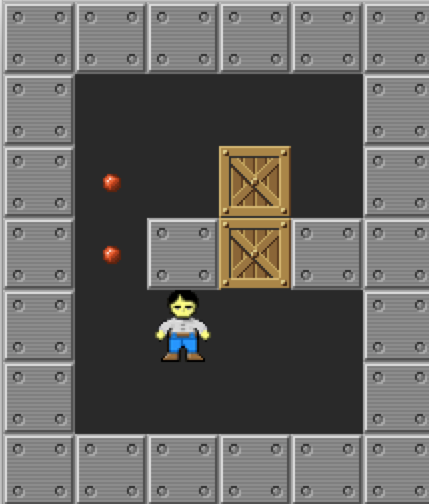}
		\end{subfigure}
	\begin{subfigure}[h!]{1.3in}
		\includegraphics[height=1.5in]{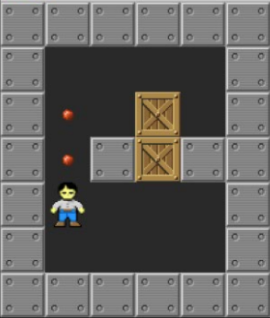}
\end{subfigure}
\begin{subfigure}[h!]{1.3in}
	\includegraphics[height=1.5in]{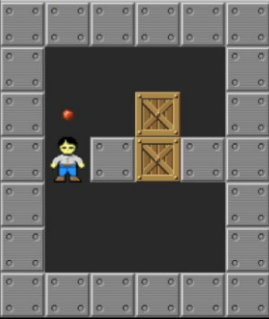}
\end{subfigure}}
		\label{fig:chain}
	\end{figure}

			\begin{figure}[h!]
		{
			\begin{subfigure}[h!]{1.3in}
				\includegraphics[height=1.5in]{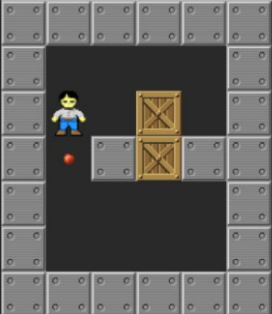}
			\end{subfigure}
			\begin{subfigure}[h!]{1.3in}
				\includegraphics[height=1.5in]{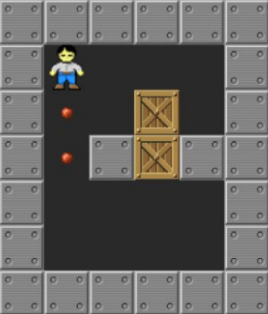}
			\end{subfigure}
			\begin{subfigure}[h!]{1.3in}
				\includegraphics[height=1.5in]{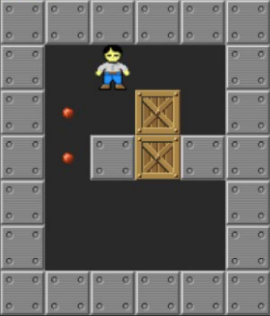}
			\end{subfigure}
			\begin{subfigure}[h!]{1.3in}
				\includegraphics[height=1.5in]{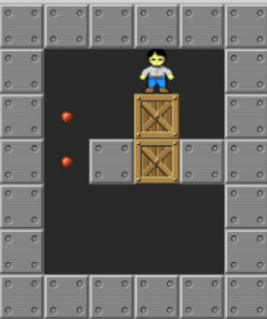}
			\end{subfigure}
			\begin{subfigure}[h!]{1.3in}
				\includegraphics[height=1.5in]{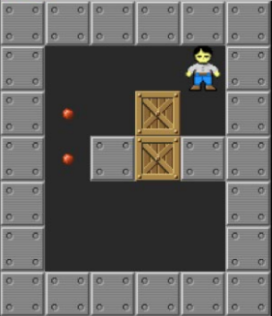}
		\end{subfigure}}
		\label{fig:chain}
	\end{figure}
				\begin{figure}[h!]
			{
				\begin{subfigure}[h!]{1.3in}
					\includegraphics[height=1.5in]{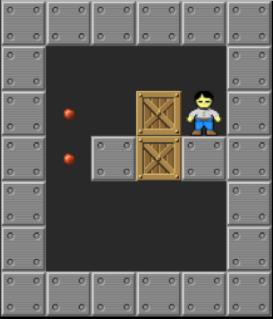}
				\end{subfigure}
				\begin{subfigure}[h!]{1.3in}
					\includegraphics[height=1.5in]{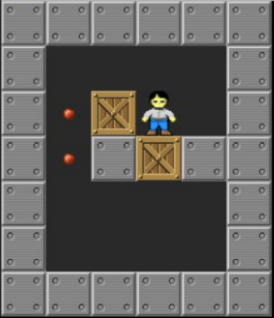}
				\end{subfigure}
				\begin{subfigure}[h!]{1.3in}
					\includegraphics[height=1.5in]{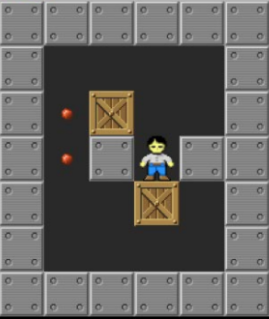}
				\end{subfigure}
				\begin{subfigure}[h!]{1.3in}
					\includegraphics[height=1.5in]{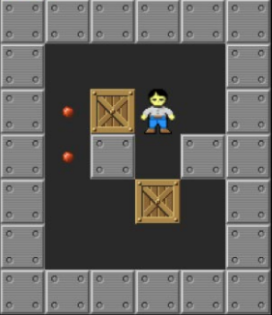}
				\end{subfigure}
				\begin{subfigure}[h!]{1.3in}
					\includegraphics[height=1.5in]{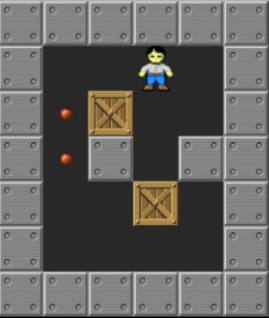}
			\end{subfigure}}
			\label{fig:chain}
		\end{figure}
		
		\begin{figure}[h!]
			{
				\begin{subfigure}[h!]{1.3in}
					\includegraphics[height=1.5in]{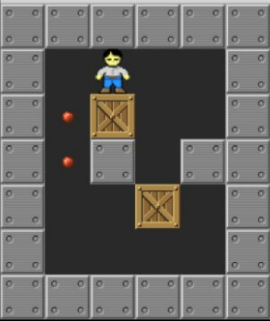}
				\end{subfigure}
				\begin{subfigure}[h!]{1.3in}
					\includegraphics[height=1.5in]{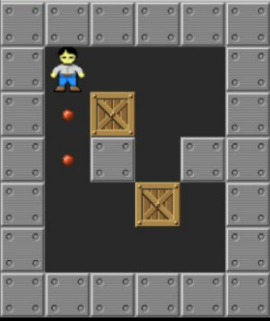}
				\end{subfigure}
				\begin{subfigure}[h!]{1.3in}
					\includegraphics[height=1.5in]{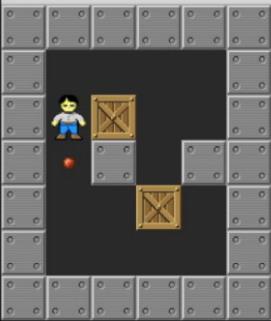}
				\end{subfigure}
				\begin{subfigure}[h!]{1.3in}
					\includegraphics[height=1.5in]{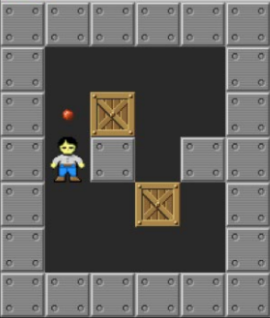}
				\end{subfigure}
				\begin{subfigure}[h!]{1.3in}
					\includegraphics[height=1.5in]{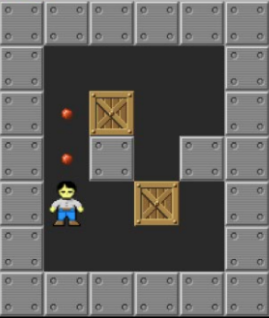}
			\end{subfigure}}
			\label{fig:chain}
		\end{figure}
				\begin{figure}[h!]
			{
				\begin{subfigure}[h!]{1.3in}
					\includegraphics[height=1.5in]{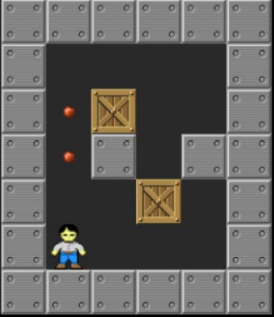}
				\end{subfigure}
				\begin{subfigure}[h!]{1.3in}
					\includegraphics[height=1.5in]{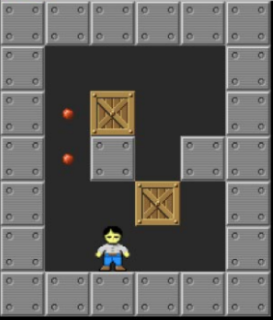}
				\end{subfigure}
				\begin{subfigure}[h!]{1.3in}
					\includegraphics[height=1.5in]{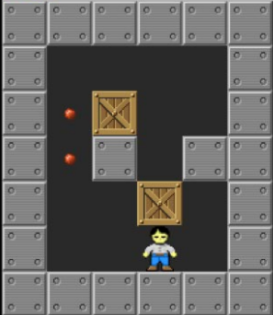}
				\end{subfigure}
				\begin{subfigure}[h!]{1.3in}
					\includegraphics[height=1.5in]{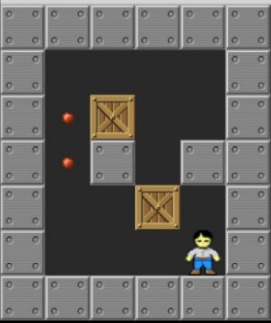}
				\end{subfigure}
				\begin{subfigure}[h!]{1.3in}
					\includegraphics[height=1.5in]{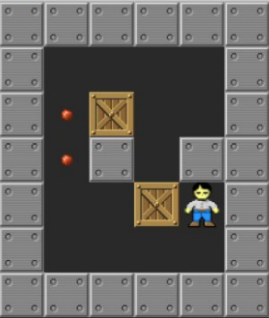}
			\end{subfigure}}
			\label{fig:chain}
		\end{figure}
		
		\begin{figure}[h!]
			{
				\begin{subfigure}[h!]{1.3in}
					\includegraphics[height=1.5in]{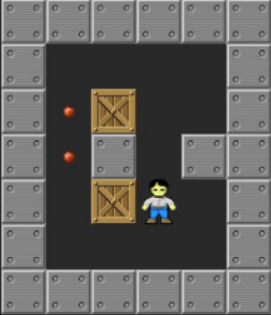}
				\end{subfigure}
				\begin{subfigure}[h!]{1.3in}
					\includegraphics[height=1.5in]{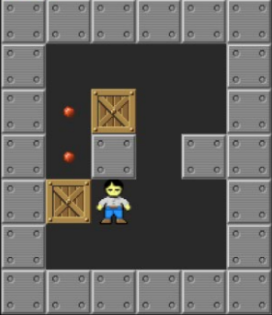}
				\end{subfigure}
				\begin{subfigure}[h!]{1.3in}
					\includegraphics[height=1.5in]{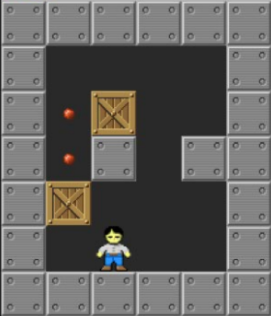}
				\end{subfigure}
				\begin{subfigure}[h!]{1.3in}
					\includegraphics[height=1.5in]{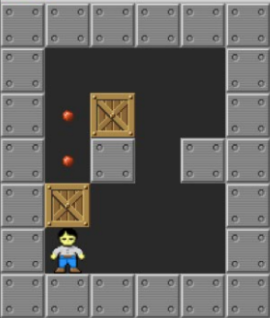}
				\end{subfigure}
				\begin{subfigure}[h!]{1.3in}
					\includegraphics[height=1.5in]{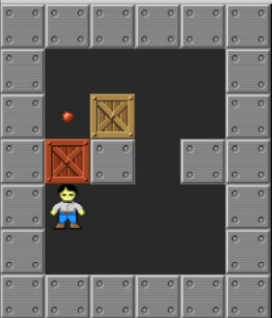}
			\end{subfigure}}
			\label{fig:chain}
		\end{figure}
						\begin{figure}[h!]
			{
				\begin{subfigure}[h!]{1.3in}
					\includegraphics[height=1.5in]{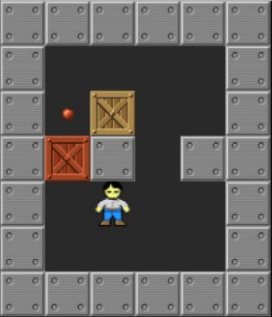}
				\end{subfigure}
				\begin{subfigure}[h!]{1.3in}
					\includegraphics[height=1.5in]{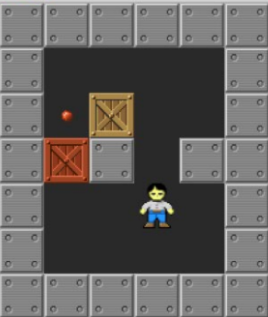}
				\end{subfigure}
				\begin{subfigure}[h!]{1.3in}
					\includegraphics[height=1.5in]{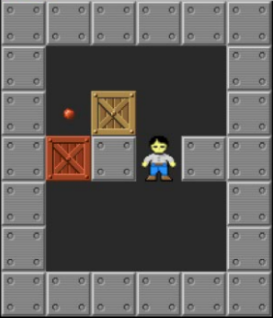}
				\end{subfigure}
				\begin{subfigure}[h!]{1.3in}
					\includegraphics[height=1.5in]{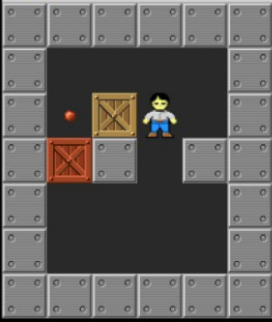}
				\end{subfigure}
				\begin{subfigure}[h!]{1.3in}
					\includegraphics[height=1.5in]{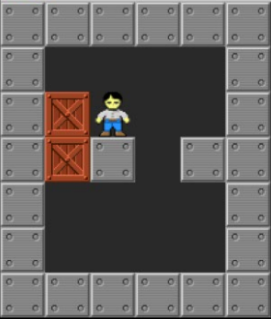}
			\end{subfigure}}
			\label{fig:chain}
		\end{figure}

\end{document}